\documentclass[runningheads]{llncs}
\usepackage[T1]{fontenc}
\usepackage{graphicx,verbatim}
\usepackage{booktabs}

\usepackage{amssymb}
\usepackage{amsmath}
\usepackage{siunitx}
\begin{document}
\title{XOCT: Enhancing OCT to OCTA Translation via Cross-Dimensional Supervised Multi-Scale Feature Learning}
\titlerunning{XOCT: OCT to OCTA via Cross-Dimensional Multi-Scale Learning}

\author{
Pooya Khosravi\textsuperscript{*}\textsuperscript{\textdagger} \and
Kun Han\textsuperscript{*} \and
Anthony T. Wu \and
Arghavan Rezvani \and
Zexin Feng \and
Xiaohui Xie
}
% index {Khosravi, Pooya}
% index {Han, Kun}
% index {Wu, Anthony T.}
% index {Rezvani, Arghavan}
% index {Feng, Zexin}
% index {Xie, Xiaohui}

\authorrunning{P. Khosravi et al.}
\institute{University of California, Irvine, CA, USA}

\maketitle              % typeset the header of the contribution
\begin{abstract}
Optical Coherence Tomography Angiography (OCTA) and its derived en-face projections provide high-resolution visualization of the retinal and choroidal vasculature, which is critical for the rapid and accurate diagnosis of retinal diseases. However, acquiring high-quality OCTA images is challenging due to motion sensitivity and the high costs associated with software modifications for conventional OCT devices. Moreover, current deep learning methods for OCT-to-OCTA translation often overlook the vascular differences across retinal layers and struggle to reconstruct the intricate, dense vascular details necessary for reliable diagnosis.
To overcome these limitations, we propose XOCT, a novel deep learning framework that integrates Cross-Dimensional Supervision (CDS) with a Multi-Scale Feature Fusion (MSFF) network for layer-aware vascular reconstruction. 
Our CDS module leverages 2D layer-wise en-face projections, generated via segmentation-weighted z-axis averaging, as supervisory signals to compel the network to learn distinct representations for each retinal layer through fine-grained, targeted guidance.
Meanwhile, the MSFF module enhances vessel delineation through multi-scale feature extraction combined with a channel reweighting strategy, effectively capturing vascular details at multiple spatial scales.  
Our experiments on the OCTA-500 dataset demonstrate XOCT's improvements, especially for the en-face projections which are significant for clinical evaluation of retinal pathologies, underscoring its potential to enhance OCTA accessibility, reliability, and diagnostic value for ophthalmic disease detection and monitoring. The code is available at https://github.com/uci-cbcl/XOCT.

% Authors must provide keywords and are not allowed to remove this Keyword section.

\end{abstract}

\renewcommand{\thefootnote}{\fnsymbol{footnote}}
\setcounter{footnote}{0}

\footnotetext{\textsuperscript{*} These authors contributed equally to this work.}
\footnotetext{\textsuperscript{\textdagger} Corresponding author.}

\section{Introduction}
\begin{figure}
\label{fig1}
\includegraphics[width=\textwidth]{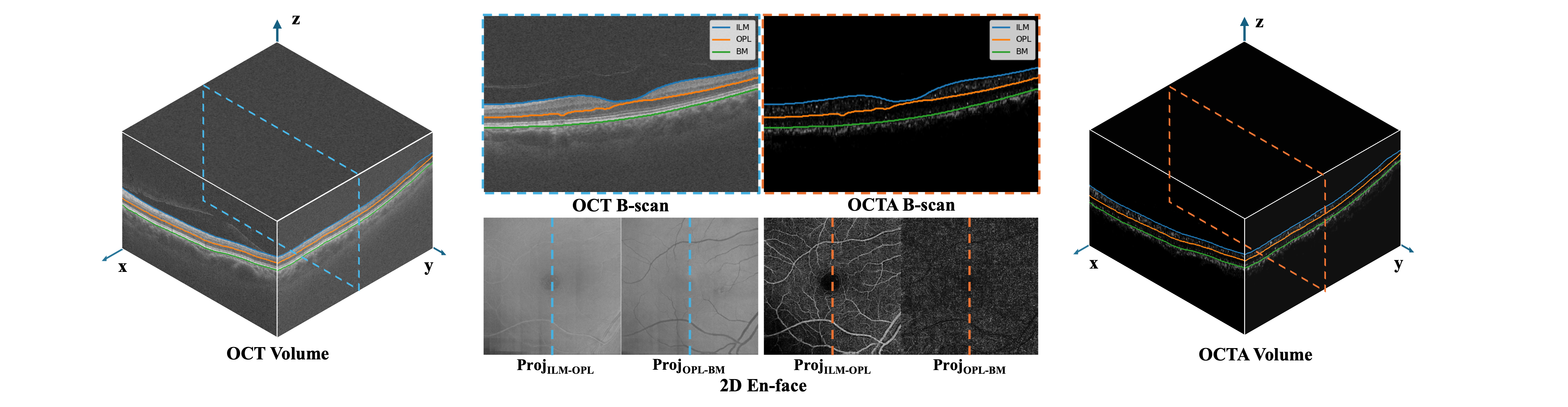}
\caption{OCT/OCTA volumes with retinal layer segmentation and en-face projections. The differences between $\text{Proj}_{\text{ILM-OPL}}$ and $\text{Proj}_{\text{OPL-BM}}$ highlight heterogeneous imaging properties due to different cellular and vascular distributions across retinal layers.}
\end{figure}

Optical Coherence Tomography Angiography (OCTA) has transformed retinal imaging by enabling high-resolution, dye-free visualization of retinal and choroidal microvasculature. The en-face OCTA projections shown in Fig. \ref{fig1}, provide intuitive, top-down views of the vascular network, facilitating rapid assessment of retinal pathologies. This non-invasive modality is crucial for early detection and monitoring of conditions such as diabetic retinopathy, age-related macular degeneration (AMD), and glaucoma \cite{kashani_optical_2017}.
However, acquiring high-quality OCTA images remains challenging due to motion artifacts and the high costs of software modifications required for OCTA-enabled devices \cite{hormet_octa_artifact_2021,de2019controlling,song2019first,anvari2021artifacts}. 

Recent deep learning–based OCT-to-OCTA translation methods show promise but face key limitations for clinical adoption. 2D B-scan-based approaches  \cite{grfm_2d,tgu_2d,dloct_2d,dlmas_2d,wsdl_2d,dlp_2d} fail to preserve 3D vascular continuity, leading to fragmented reconstructions that compromise network integrity. Projection-based methods \cite{multi-gan} reduce volumetric data to a single 2D plane (Fig. \ref{fig1}), obscuring fine vascular details and reducing angiogram fidelity. Meanwhile, 3D-based models \cite{pboct_3d,transpro} rely on conventional feature extraction, failing to leverage layer-specific retinal properties needed to capture subtle microvascular structures for clinical interpretation.

To address these challenges, we propose \textbf{XOCT}, a novel framework that combines 2D and 3D insights to preserve fine vascular details across heterogeneous retinal layers. XOCT integrates two key components: Cross-Dimensional Supervision (\textbf{CDS}) and a Multi-Scale Feature Fusion (\textbf{MSFF}) network.

The CDS module leverages the heterogeneous imaging properties of retinal layers by integrating volumetric and layer-wise constraints. As shown in Fig. \ref{fig1}, variations in tissue composition and vascular distribution cause each layer to interact differently with light, producing unique structural patterns \cite{campbell_vascular_retina_octa}. CDS generates 2D en-face projections via segmentation-weighted z-axis averaging, aligning them with ground-truth maps using a composite loss: $L_1$ loss for pixel-wise accuracy, adversarial loss for anatomical realism, and perceptual loss \cite{Perp} for high-level structural fidelity. By providing fine-grained, layer-specific supervision, CDS encourages the network to learn distinct feature representations for each retinal layer, enforcing intra-layer consistency, preserving vessel coherence, and capturing subtle microvascular details that conventional methods often miss.

The \textbf{MSFF} module is designed to refine vessel delineation by capturing vascular details across multiple spatial scales. Recognizing that OCTA images feature extremely thin and intricate vascular structures, MSFF employs a combination of isotropic kernels for balanced local feature extraction and anisotropic kernels tailored to detect the elongated patterns of retinal vessels. Additionally, depth-wise large-kernel convolutions are incorporated to broaden the receptive field, ensuring that global vessel connectivity is effectively captured. To optimize computational efficiency, the output channels of each convolutional block are halved, and the resulting multi-scale features are subsequently fused via point-wise convolution coupled with a channel reweighting mechanism. This adaptive fusion process emphasizes critical vascular details, thereby enhancing the overall fidelity of OCT-to-OCTA translation by preserving both fine local structures and the broader vascular network.

The main contributions of our work are: 
(1) \textbf{XOCT}, a deep learning framework that integrates \textbf{CDS} and \textbf{MSFF} for OCT-to-OCTA translation.
(2) \textbf{CDS}, the first method to incorporate retinal layer characteristics during training, preserving vessel coherence and structural integrity.
(3) \textbf{MSFF}, an efficient module that enhances fine vascular detail capture via multi-scale feature fusion.
(4) Extensive evaluation on the OCTA-500 dataset \cite{octa-500}, demonstrating superior vascular clarity, continuity, and translation performance in both 3D OCTA volumes and layer-wise en-face projections with direct clinical relevance.

\section{Related Works}

\textbf{2D B-scan-Based Approaches}: Early OCT-to-OCTA translation methods focused on individual 2D B-scans.
Lee et al. \cite{grfm_2d} proposed an encoder-decoder-based framework for mapping paired OCT B-scans to OCTA images.
Zhang et al. \cite{tgu_2d} improved vascular detail preservation with texture-guided down-sampling, while Li et al. \cite{dloct_2d} incorporated adversarial loss to enhance image fidelity. Despite these advancements, the lack of volumetric modeling limited their ability to accurately reconstruct retinal vasculature.

\textbf{Projection-Based Approaches}: These methods convert OCT and OCTA volumes into 2D en face representations before translation. Pan et al. \cite{multi-gan} proposed MultiGAN, an unsupervised multi-domain framework that generated OCTA projection maps from OCT projections, enforcing anatomical consistency through domain-specific loss functions. 
However, the projection process inherently discards depth information, leading to the loss of fine vascular structures.

\textbf{3D-Based Approaches}: Recent studies have adopted volumetric OCT-to-OCTA translation to address 2D-based limitations. Huang et al. \cite{pboct_3d} introduced a patch-based 3D model with a context-enhanced encoder, while Li et al. \cite{transpro} developed TransPro, a 3D Pix2Pix framework refined via supervision from pretrained 2D models. However, these methods overlook retinal layer-specific imaging characteristics and depend on standard convolutional operations, which struggle to capture intricate vessel structures present in OCTA images.

\begin{figure}
\includegraphics[width=\textwidth]{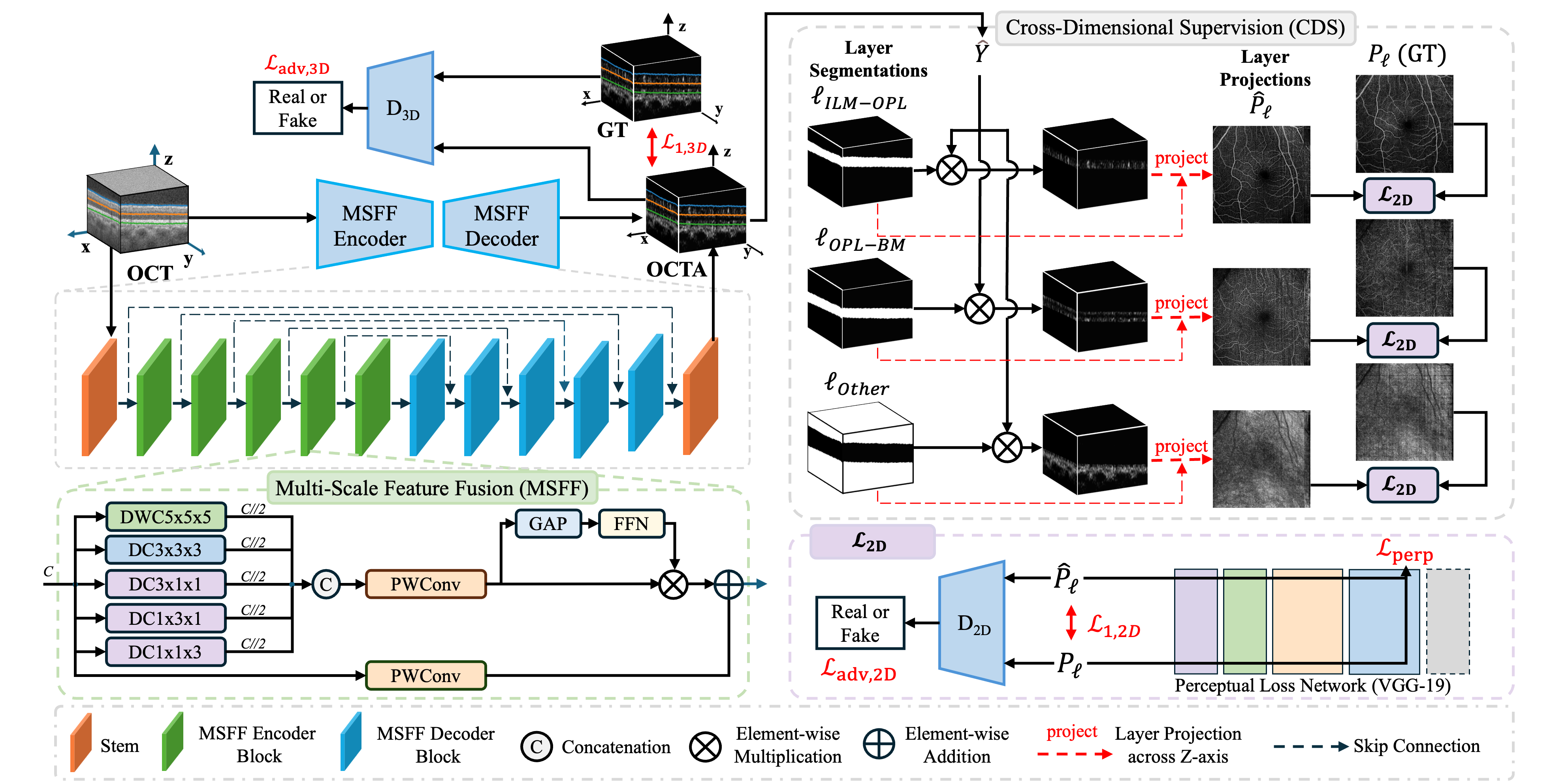}
\caption{Overview of XOCT.} \label{fig:xoct}
\end{figure}

\section{Methods}
We propose XOCT (Fig. \ref{fig:xoct}), a novel deep learning framework for OCT-to-OCTA translation that builds on a 3D encoder-decoder architecture enhanced by two key components: Cross-Dimensional Supervision (CDS) and a Multi-Scale Feature Fusion (MSFF) module. XOCT accepts a volumetric OCT scan $\mathbf{X} \in \mathbb{R}^{D \times H \times W}$ as input and outputs a reconstructed OCTA volume $\widehat{\mathbf{Y}} \in \mathbb{R}^{D \times H \times W}$. We then implement an end-to-end training with a composite loss function combining volumetric and projection-based objectives to ensure both global structural consistency and precise vessel delineation.

\subsection{Cross-Dimensional Supervision}

Conventional volumetric supervision treats the OCTA volume as homogeneous, overlooking the retina's intrinsic heterogeneity. In reality, the retina consists of multiple layers, each with distinct tissue compositions, cellular structures, and vascular distributions \cite{campbell_vascular_retina_octa}. To capture these nuances and preserve fine vascular details, we propose a Cross-Dimensional Supervision (CDS), which augments standard 3D supervision with targeted layer-wise guidance.

Given a retinal layer segmentation map $\mathbf{S} \in \mathbb{R}^{D \times H \times W}$, we generate 2D layer-specific projection maps $\mathbf{P}_l \in \mathbb{R}^{H \times W}$ for each layer $l$. For a predicted OCTA volume $\widehat{\mathbf{Y}}$, the corresponding layer-specific projection is computed as the segmentation-weighted average of voxel intensities along the \textbf{z-axis}:

\begin{equation}\label{proj_2d}
\widehat{\mathbf{P}_l} = \frac{\sum_z \widehat{\mathbf{Y}} \odot \mathbf{S}_l}{\sum_{z} \mathbf{S}_l},
\end{equation}
where $\odot$ denotes element-wise multiplication. 

The predicted projections, $\widehat{\mathbf{P}_l}$, are compared to their ground truth counterparts, $\mathbf{P}_l$, using a composite loss function:
\begin{equation}
L_{2D} = \sum_l \Big( \alpha_{2D}L_{1}(\widehat{\mathbf{P}_l}, \mathbf{P}_l) + \beta_{2D} L_{adv}(\widehat{\mathbf{P}_l}, \mathbf{P}_l) + \gamma_{2D} L_{perp}(\widehat{\mathbf{P}_l}, \mathbf{P}_l) \Big),
\end{equation}

\noindent where $L_{1}$ minimizes pixel-wise differences, $L_{adv}$ promotes realism through adversarial training, and $L_{perp}$—based on a pretrained VGG19 network—ensures high-level perceptual fidelity. Supervising each layer individually encourages the network to learn distinct representations for different layers, preserve intra-layer consistency, and accurately reconstruct vascular details specific to each layer. 

Integrating this layer-aware supervision with volumetric loss: $L = L_{3D} + L_{2D}$, 
enables the network to preserve vessel continuity and structural integrity across retinal layers, enhancing the overall fidelity of OCT-to-OCTA translation.

\subsection{Multi-Scale Feature Fusion}

We introduce the Multi-Scale Feature Fusion (MSFF) module, which integrates local and broader context across spatial scales for enhanced fine vasculature reconstruction. MSFF employs both \textbf{isotropic} and \textbf{anisotropic} convolution kernels. Isotropic 3×3×3 convolutions capture balanced spatial information, while anisotropic kernels (3×1×1, 1×3×1, and 1×1×3) extract elongated vessel features. Additionally, \textbf{depth-wise} large-kernel (5x5x5) convolutions are used to expand the receptive field and capture broader vessel connectivity. Although larger kernels (e.g., 7×7×7) offered only marginal performance gains, the 5×5×5 configuration was selected as a trade-off between accuracy and computational efficiency, avoiding the cubic scaling cost of larger kernels. This architecture is specifically optimized for 3D vascular reconstruction and contrasts with prior multi-scale segmentation methods that employ varying kernel shapes and configurations \cite{multi-scale}.

To enhance efficiency while maintaining performance, we halved the output channels of each convolutional block, reducing parameter count. Multi-scale features are fused via point-wise convolution and channel re-weighting, adaptively emphasizing critical vascular details across different spatial scales. A \textbf{residual connection} from the module’s input further preserves low-level details and facilitates gradient flow. This multi-scale strategy effectively delineates delicate vasculature in OCT and OCTA, improving vascular reconstruction fidelity.

\subsection{Overall Framework}

During training, we jointly optimize the 3D volumetric generator $G_{3D}$, the volumetric discriminator $D_{3D}$ and the 2D projection discriminators $D^l_{2D}$ from different retinal layers $l$ under a generative adversarial learning paradigm \cite{pix2pix}.

The overall volumetric loss is defined as:
\begin{equation}
L_{3D} = \alpha_{3D} L_{1}(\widehat{\mathbf{Y}}, \mathbf{Y}) + \beta_{3D} L_{adv}(\widehat{\mathbf{Y}}, \mathbf{Y}),
\end{equation}
where $L_{1}$ minimizes the pixel-wise differences between the predicted OCTA volume $\widehat{\mathbf{Y}}$ and the ground truth $\mathbf{Y}$, and $L_{adv}$ is the adversarial loss given by:
\begin{equation}
L_{adv} = \mathbb{E}_{\mathbf{Y} \sim p}\Big[\log \big(D(\mathbf{Y})\big)\Big] + \mathbb{E}_{\mathbf{X} \sim p}\Big[\log \big(1-D\big(G(\mathbf{X})\big)\big)\Big].
\end{equation}

Note that for 2D supervision, en-faceprojections are directly generated from the predicted OCTA volume using retinal layer segmentation maps, eliminating the need for a separate 2D generator (Eq. \ref{proj_2d}). These projections are evaluated using an adversarial framework with additional $L_1$  loss and  perceptual loss \cite{Perp}. The segmentation maps are used exclusively during training for 2D projection supervision and are not required for OCT-to-OCTA translation during inference.

\section{Experiments}

\subsection{Datasets and Experimental Setup} 

\textbf{Dataset:} The publicly available \textbf{OCTA-500} \cite{octa-500} comprises of 500 3D OCT-OCTA volume pairs and supplementary annotations, including retinal layer segmentation. OCTA-500 is divided into two subsets: \textbf{OCTA-3M} and \textbf{OCTA-6M}. The \textbf{OCTA-3M} contains 200 scans with a field-of-view of $3mm\times3mm\times2mm$ and a volume size of $304\times304\times640$ pixels, partitioned into 140 training, 20 validation, and 40 test scans. The \textbf{OCTA-6M} consists of 300 scans with a field-of-view of $6mm\times6mm\times2mm$ and a volume size of $400\times400\times640$ pixels, divided into 200 training, 30 validation, and 70 test scans.

\noindent \textbf{Implementation Details:} 
XOCT utilizes a modified 3D Pix2Pix architecture trained for 300 epochs with Adam optimizer, a learning rate of \num{1e-4}, and a batch size of 1. The adversarial loss weights were fixed at 1, while a grid search determined the optimal $L_1$ and perceptual loss weights to be $\lambda_{L_1}=10$ and $\lambda_{perp}=1$, respectively.

\noindent \textbf{Experimental Setup:}
We evaluate the performance of XOCT by assessing both the en-face projections and reconstructed 3D volumes. For the en-face evaluation, we use projections that differ from those employed during training to validate our contributions. Specifically, $\textbf{Proj}_{\textbf{full}}$ spans from the internal limiting membrane (ILM) to Bruch’s membrane (BM), capturing the complete vascular structure, while $\textbf{Proj}_{\textbf{mean}}$ is computed as the mean projection across the entire z-axis, offering a representative view of the overall vasculature. Evaluation metrics include Mean Absolute Error (MAE), Peak Signal-to-Noise Ratio (PSNR)\cite{PSNR}, Structural Similarity (SSIM)\cite{SSIM}, and Perceptual Discrepancy (Perp.)\cite{Perp}.

We compare XOCT with leading 2D-based methods (BBDM \cite{bbdm}, Pix2Pix \cite{pix2pix}, MultiGAN \cite{multi-gan}) and 3D-based methods (BBDM3D* \cite{bbdm3d}, Pix2Pix3D \cite{pix2pix3d}, TransPro \cite{transpro}), demonstrating the improvements achieved by our model.  Recent studies, such as \cite{bbdm3d,pasta}, address the high GPU memory consumption of 3D diffusion by adopting a 2.5D strategy that stacks multiple neighboring slices as a single input. In our experiments, we use BBDM3D* \cite{bbdm3d} as a representative diffusion method for comparison.

\subsection{Experiment Results}

% Numbers need to be revised, just sample table formatting.
\begin{table*}[h]
\centering
\caption{OCT to OCTA translation results on the \textbf{OCTA-3M} and \textbf{OCTA-6M} datasets. $\downarrow$ indicates lower is better, $\uparrow$ indicates higher is better.}
\label{tab:main_table}
\resizebox{\textwidth}{!}{
\begin{tabular}{l|cccc|cccc|cccc}
\hline
\textbf{OCTA-3M} & \multicolumn{4}{c|}{$\text{Proj}_{\text{full}}$} & \multicolumn{4}{c|}{$\text{Proj}_{\text{mean}}$} & \multicolumn{4}{c}{3D}  \\
\hline
Method & MAE $\downarrow$ & PSNR $\uparrow$ & SSIM $\uparrow$ & Perp. $\downarrow$ & MAE $\downarrow$ & PSNR $\uparrow$ & SSIM $\uparrow$  & Perp. $\downarrow$ & MAE $\downarrow$ & PSNR $\uparrow$ & SSIM $\uparrow$  & Perp. $\downarrow$ \\
\hline
BBDM \cite{bbdm} & 22.86 & 18.84 & 0.420 & 0.741 & 25.33 & 18.14 & 0.384 & 0.727 & - & - & - & - \\
Pix2Pix \cite{pix2pix} & 24.25 & 18.16 & 0.380 & 0.726 & 27.18 & 17.47 & 0.340 & 0.708 & - & - & - & -  \\
MultiGAN \cite{multi-gan} & 22.54 & 18.87 & 0.417 & 0.722 & 24.73 & 18.32 & 0.382 & 0.708 & - & - & - & -  \\
\hline
BBDM3D* \cite{bbdm3d} & 21.04 & 19.54 & 0.460 & 0.703 & 22.51 & 19.27 & 0.457 & 0.675 & \textbf{2.69} & \textbf{30.71} & 0.883  & 0.202 \\
Pix2Pix3D \cite{pix2pix3d} & 19.87 & 19.91 & 0.556 & 0.656 & 21.61 & 19.47 & 0.541 & 0.620 & 2.88 & 29.45 & 0.885 & 0.198 \\
TransPro \cite{transpro} & 19.54 & 20.14 & 0.580 & 0.608 & 20.93 & 19.83 & 0.573 & 0.582 & 3.30\textsuperscript{\textdagger} & 28.46 & 0.866 & 0.203 \\
\hline
XOCT & \textbf{19.22} & \textbf{20.21} & \textbf{0.608} & \textbf{0.573} & \textbf{20.54} & \textbf{19.90} & \textbf{0.596} & \textbf{0.558} & 2.78 & 30.08 & \textbf{0.893} & \textbf{0.184} \\
\hline

\multicolumn{10}{c}{} \\

\hline
\textbf{OCTA-6M} & \multicolumn{4}{c|}{$\text{Proj}_{\text{full}}$} & \multicolumn{4}{c|}{$\text{Proj}_{\text{mean}}$} & \multicolumn{4}{c}{3D}  \\
\hline
Method & MAE $\downarrow$ & PSNR $\uparrow$ & SSIM $\uparrow$ & Perp. $\downarrow$ & MAE $\downarrow$ & PSNR $\uparrow$ & SSIM $\uparrow$  & Perp. $\downarrow$ & MAE $\downarrow$ & PSNR $\uparrow$ & SSIM $\uparrow$  & Perp. $\downarrow$ \\
\hline
BBDM \cite{bbdm} & 22.94 & 18.68 & 0.367 & 0.729 & 25.73 & 17.92 & 0.346 & 0.714 & - & - & - & - \\
Pix2Pix \cite{pix2pix} & 22.87 & 18.67 & 0.347 & 0.703 & 26.43 & 17.76 & 0.303 & 0.695 & - & - & - & - \\
MultiGAN \cite{multi-gan} & 21.22 & 19.32 & 0.374 & 0.713 & 24.69 & 18.43 & 0.332 & 0.714 & - & - & - & - \\
\hline
BBDM3D* \cite{bbdm3d} & 17.90 & 20.81 & 0.433 & 0.702 & 18.29 & 20.81 & 0.436 & 0.686 & 2.48 & \textbf{31.44} & 0.889 & 0.146 \\
Pix2Pix3D \cite{pix2pix3d} & 17.39 & 21.06 & 0.494 & 0.641 & 17.21 & 21.33 &  0.510 & 0.609 & \textbf{2.34} & 31.23 & 0.903 & 0.136 \\
TransPro \cite{transpro} & 17.40 & 21.07 & 0.524 & 0.591 & 17.42 & 21.24 & 0.539 & 0.570 & 2.72\textsuperscript{\textdagger} & 29.92 & 0.887 & 0.155 \\
\hline
XOCT & \textbf{16.65} & \textbf{21.46} & \textbf{0.568} & \textbf{0.541} & \textbf{16.21} & \textbf{21.85} & \textbf{0.574} & \textbf{0.537} & \textbf{2.34} & 31.26 & \textbf{0.905} & \textbf{0.131} \\
\hline
\end{tabular}
}
\begin{flushleft} % Using flushleft to align the note with the table and allow text wrapping
\textsuperscript{\textdagger}\scriptsize{The reported MAE in TransPro\cite{transpro} appears higher (e.g., 0.078 vs 3.30/255=0.013) due to a \texttt{uint8} overflow error (e.g., \texttt{abs}(1-2) = 255). All reported MAE values use \texttt{float32} for a fair comparison.}
\end{flushleft}
\end{table*}

Table \ref{tab:main_table} presents the quantitative results for the OCTA-3M and OCTA-6M datasets across 2D en-face projections ($\textbf{Proj}_{\textbf{full}}$, $\textbf{Proj}_{\textbf{mean}}$) and 3D OCTA volumes. 2D projection-based methods (BBDM, Pix2Pix, MultiGAN) exhibit poor reconstruction of fine vascular details due to volumetric collapse, leading to lower PSNR/SSIM and higher MAE/Perp. Although BBDM3D* improves vessel reconstruction by stacking B-scans as channels, it remains suboptimal usingvolumetric structure. Similarly, TransPro, which employs a 3D Pix2Pix framework with pretrained 2D supervision, inherits the constraints of its 2D components, resulting in compromised 3D reconstructions.

\begin{figure}
\includegraphics[width=\textwidth]{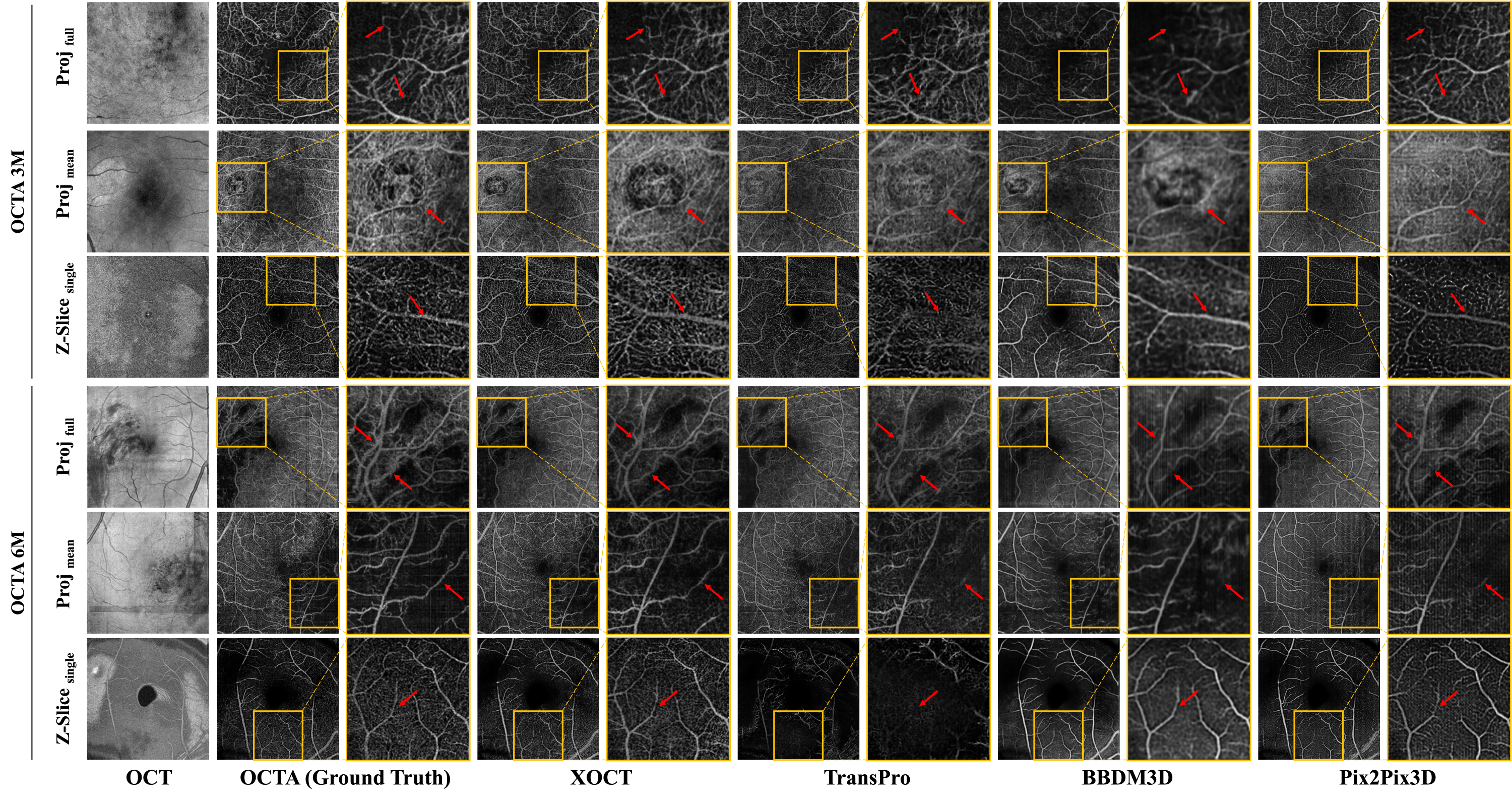}
\caption{Comparison of OCT-to-OCTA translation methods across 2D en-face projections ($\textbf{Proj}_{\textbf{full}}$, $\textbf{Proj}_{\textbf{mean}}$) and a 3D z-slice cross-section. Red arrows highlight where XOCT achieves enhanced vascular connectivity, accurately generates vessels in regions where other models fail, and better preserves subtle vessel dropouts. XOCT reconstructs fine vascular structures with greater clarity and continuity, reducing artifacts and improving the delineation of small vessels that other methods struggle to resolve.} \label{fig3}
\end{figure}

Although BBDM3D* achieved lower MAE and higher PSNR in 3D evaluation, qualitative analysis (Fig. \ref{fig3}) revealed blurring, particularly in small vessels. This suggests that while BBDM3D* captures overall intensity well, reducing pixel-wise error, it fails to preserve high-frequency microvascular details, instead prioritizing coarse structural consistency. 
In contrast, XOCT produces sharper and more anatomically faithful vessels, and while this leads to superior qualitative results, metrics like 3D MAE, which are sensitive to overall intensity, may not fully capture these critical structural improvements.

Beyond quantitative metrics, XOCT excels in preserving vessel continuity and fine vascular structures, as demonstrated in Fig. \ref{fig3}. It more accurately captures subtle dropouts in vessel density, which are critical in detecting microvascular abnormalities associated with diseases such as diabetic retinopathy \cite{diabetic_vessel_loss}. In contrast, Pix2Pix3D and TransPro tend to overcompensate by filling in missing vessel regions, potentially obscuring clinically significant perfusion deficits. XOCT’s ability to maintain vessel continuity across varying scales underscores its robustness in generating clinically meaningful OCTA reconstructions.

\subsection{Ablation Study}
Table \ref{tab:ablation} presents an ablation study that evaluates the proposed CDS and MSFF modules, highlighting significant improvements in both components. Specifically, CDS enhances en-face projection metrics by incorporating layer-aware 2D supervision, reinforcing vascular integrity retinal layer-specific constraints. Meanwhile, MSFF improves 3D reconstruction by capturing fine volumetric details via multi-scale feature extraction with less parameters. When integrated, CDS and MSFF enable XOCT to preserve vascular coherence and maintain detailed volumetric information, outperforming the baseline across all metrics.

%%%%% Original Table - commented to make smaller
\begin{table*}[h]
\centering
\caption{Ablation Study on Proposed Modules. $\downarrow$: lower is better, $\uparrow$: higher is better.}
\label{tab:ablation}
\resizebox{\textwidth}{!}{
\begin{tabular}{l|c|cc|cc|cc|cc|cc|}
\hline
\textbf{OCTA-3M} & \# Param. & \multicolumn{2}{c|}{$\text{Proj}_{\text{full}}$} & \multicolumn{2}{c|}{$\text{Proj}_{\text{ILM-OPL}}$} & \multicolumn{2}{c|}{$\text{Proj}_{\text{OPL-BM}}$} & \multicolumn{2}{c|}{$\text{Proj}_{\text{mean}}$} & \multicolumn{2}{c|}{3D}\\
\hline
Method & & SSIM $\uparrow$ & Perp. $\downarrow$ & SSIM $\uparrow$ & Perp. $\downarrow$  & SSIM $\uparrow$ & Perp. $\downarrow$  & SSIM $\uparrow$ & Perp. $\downarrow$  & SSIM $\uparrow$ & Perp. $\downarrow$  \\
\hline
Pix2Pix3D & 64.6M & 0.556 & 0.656 & 0.509 & 0.720 & 0.479 & 0.668 & 0.541 & 0.620 & 0.885 & 0.198\\
\hline
+CDS & 64.6M & 0.600 & 0.581 & 0.563 & 0.606 & 0.522 & 0.616 & 0.590 & 0.563 & 0.889 & 0.195\\
+MSFF & \textbf{52.7M} & 0.589 & 0.651 & 0.552 & 0.707 & 0.523 & 0.677 & 0.577 & 0.614 & 0.893 & 0.184\\
\hline
XOCT & \textbf{52.7M} & \textbf{0.608} & \textbf{0.573} & \textbf{0.577} & \textbf{0.592} & \textbf{0.533} & \textbf{0.609} & \textbf{0.596} & \textbf{0.558} & \textbf{0.893} & \textbf{0.184}\\
\hline

\multicolumn{10}{c}{} \\

\hline
\textbf{OCTA-6M} & \# Param. & \multicolumn{2}{c|}{$\text{Proj}_{\text{full}}$} & \multicolumn{2}{c|}{$\text{Proj}_{\text{ILM-OPL}}$} & \multicolumn{2}{c|}{$\text{Proj}_{\text{OPL-BM}}$} & \multicolumn{2}{c|}{$\text{Proj}_{\text{mean}}$} & \multicolumn{2}{c|}{3D}\\
\hline
Method & & SSIM $\uparrow$ & Perp. $\downarrow$ & SSIM $\uparrow$ & Perp. $\downarrow$  & SSIM $\uparrow$ & Perp. $\downarrow$  & SSIM $\uparrow$ & Perp. $\downarrow$  & SSIM $\uparrow$ & Perp. $\downarrow$  \\
\hline
Pix2Pix3D & 64.6M & 0.494 & 0.641 & 0.485 & 0.655 & 0.419 & 0.632 & 0.510 & 0.609 & 0.903 & 0.136\\
\hline
+CDS & 64.6M & 0.560 & 0.552 & 0.532 & 0.556 & 0.494 & 0.562 & 0.567 & 0.542 & 0.902 & 0.135\\
+MSFF & \textbf{52.7M} & 0.550 & 0.636 & 0.535 & 0.643 & 0.467 & 0.641 & 0.554 & 0.607 & \textbf{0.909} & 0.132\\
\hline
XOCT & \textbf{52.7M} & \textbf{0.568} & \textbf{0.541} & \textbf{0.544} & \textbf{0.540} & \textbf{0.501} & \textbf{0.545} & \textbf{0.574} & \textbf{0.537} & 0.905 & \textbf{0.131}\\
\hline

\end{tabular}
}
\end{table*}

\section{Conclusion}
We introduced XOCT for OCT-to-OCTA translation, integrating CDS for layer-specific feature extraction and MSFF for vascular reconstruction. Experiments on OCTA-500 highlight XOCT’s improvements, particularly in 2D en-face projections, enhancing vascular continuity and microvascular detail preservation, essential for retinal pathology assessment. Despite these advancements, XOCT requires validation on clinical datasets and diverse imaging devices for robustness. Future work will enhance domain generalization, small vessel reconstruction, and computational efficiency for real-time use. Expanding XOCT to multi-modal retinal imaging could further improve its clinical applicability.

\begin{credits}
\subsubsection{\ackname} This work was supported by the National Eye Institute, NIH, under Award F30EY036725 (P.K.).

\subsubsection{\discintname}
The authors have no competing interests to declare that are relevant to the content of this article. 
\end{credits}

\bibliographystyle{splncs04}
\bibliography{references}

\begin{thebibliography}{10}
\providecommand{\url}[1]{\texttt{#1}}
\providecommand{\urlprefix}{URL }
\providecommand{\doi}[1]{https://doi.org/#1}

\bibitem{anvari2021artifacts}
Anvari, P., Ashrafkhorasani, M., Habibi, A., Falavarjani, K.G.: Artifacts in optical coherence tomography angiography. Journal of ophthalmic \& vision research  \textbf{16}(2), ~271 (2021)

\bibitem{campbell_vascular_retina_octa}
Campbell, J., Zhang, M., Hwang, T., Bailey, S., Wilson, D., Jia, Y., Huang, D.: Detailed vascular anatomy of the human retina by projection-resolved optical coherence tomography angiography. Scientific reports  \textbf{7},  42201 (2 2017). \doi{10.1038/srep42201}

\bibitem{bbdm3d}
Choo, K., Jun, Y., Yun, M., Hwang, S.J.: Slice-consistent 3d volumetric brain ct-to-mri translation with 2d brownian bridge diffusion model. In: International Conference on Medical Image Computing and Computer-Assisted Intervention. pp. 657--667. Springer (2024)

\bibitem{de2019controlling}
De~Pretto, L.R., Moult, E.M., Alibhai, A.Y., Carrasco-Zevallos, O.M., Chen, S., Lee, B., Witkin, A.J., Baumal, C.R., Reichel, E., de~Freitas, A.Z., et~al.: Controlling for artifacts in widefield optical coherence tomography angiography measurements of non-perfusion area. Scientific reports  \textbf{9}(1), ~9096 (2019)

\bibitem{pix2pix3d}
Deng, K., Yang, G., Ramanan, D., Zhu, J.Y.: 3d-aware conditional image synthesis. In: Proceedings of the IEEE/CVF Conference on Computer Vision and Pattern Recognition. pp. 4434--4445 (2023)

\bibitem{PSNR}
Gonzalez, R.C., Woods, R.E.: Digital Image Processing. Prentice Hall, 3 edn. (2008)

\bibitem{hormet_octa_artifact_2021}
Hormel, T.T., Huang, D., Jia, Y.: Artifacts and artifact removal in optical coherence tomographic angiography. Quantitative imaging in medicine and surgery  \textbf{11}(3),  1120--1133 (2021). \doi{10.21037/qims-20-730}

\bibitem{pboct_3d}
Huang, K., Su, N., Tao, Y., Li, M., Ma, X., Ji, Z., Yuan, S., Chen, Q.: Cross-device octa generation by patch-based 3d multi-scale feature adaption. IEEE Transactions on Emerging Topics in Computational Intelligence  \textbf{8}(1),  641--653 (2 2024). \doi{10.1109/TETCI.2023.3314690}

\bibitem{pix2pix}
Isola, P., Zhu, J.Y., Zhou, T., Efros, A.A.: Image-to-image translation with conditional adversarial networks. In: Proceedings of the IEEE conference on computer vision and pattern recognition. pp. 1125--1134 (2017). \doi{10.1109/CVPR.2017.632}

\bibitem{wsdl_2d}
Jiang, Z., Huang, Z., Qiu, B., Meng, X., You, Y., Liu, X., Geng, M., Liu, G., Zhou, C., Yang, K., et~al.: Weakly supervised deep learning-based optical coherence tomography angiography. IEEE Transactions on Medical Imaging  \textbf{40}(2),  688--698 (10 2021). \doi{10.1109/TMI.2020.3035154}

\bibitem{Perp}
Johnson, J., Alahi, A., Fei-Fei, L.: Perceptual losses for real-time style transfer and super-resolution. In: Computer Vision--ECCV 2016: 14th European Conference, Amsterdam, The Netherlands, October 11-14, 2016, Proceedings, Part II 14. pp. 694--711. Springer (2016)

\bibitem{kashani_optical_2017}
Kashani, A.H., Chen, C.L., Gahm, J.K., Zheng, F., Richter, G.M., Rosenfeld, P.J., Shi, Y., Wang, R.K.: Optical coherence tomography angiography: A comprehensive review of current methods and clinical applications. Progress in retinal and eye research  \textbf{60},  66--100 (2017). \doi{10.1016/j.preteyeres.2017.07.002}

\bibitem{grfm_2d}
Lee, C.S., Tyring, A.J., Wu, Y., Xiao, S., Rokem, A.S., DeRuyter, N.P., Zhang, Q., Tufail, A., Wang, R.K., Lee, A.Y.: Generating retinal flow maps from structural optical coherence tomography with artificial intelligence. Scientific reports  \textbf{9}, ~5694 (2019)

\bibitem{bbdm}
Li, B., Xue, K., Liu, B., Lai, Y.K.: Bbdm: Image-to-image translation with brownian bridge diffusion models. In: Proceedings of the IEEE/CVF conference on computer vision and pattern Recognition. pp. 1952--1961 (2023)

\bibitem{octa-500}
Li, M., Huang, K., Xu, Q., Yang, J., Zhang, Y., Ji, Z., Xie, K., Yuan, S., Liu, Q., Chen, Q.: Octa-500: A retinal dataset for optical coherence tomography angiography study. Medical image analysis  \textbf{93},  103092 (2024). \doi{10.1016/j.media.2024.103092}

\bibitem{dloct_2d}
Li, P.L., O'Neil, C., Saberi, S., Sinder, K., Wang, K., Tan, B., Hosseinaee, Z., Bizhevat, K., Lakshminarayanan, V.: Deep learning algorithm for generating optical coherence tomography angiography (octa) maps of the retinal vasculature. In: Proc. SPIE 11511, Applications of Machine Learning 2020. pp. 39--49 (2020)

\bibitem{transpro}
Li, S., Zhang, D., Li, X., Ou, C., An, L., Xu, Y., Yang, W., Zhang, Y., Cheng, K.T.: Vessel-promoted oct to octa image translation by heuristic contextual constraints. Medical Image Analysis  \textbf{98},  103311 (2024)

\bibitem{pasta}
Li, Y., Yakushev, I., Hedderich, D.M., Wachinger, C.: Pasta: Pathology-aware mri to pet cross-modal translation with diffusion models. In: International Conference on Medical Image Computing and Computer-Assisted Intervention. pp. 529--540. Springer (2024)

\bibitem{dlmas_2d}
Lin, Z., Zhang, Q., Lan, G., Xu, J., Qin, J., An, L., Huang, Y.: Deep learning for motion artifact-suppressed octa image generation from both repeated and adjacent oct scans. Mathematics  \textbf{12}, ~446 (2024)

\bibitem{dlp_2d}
Liu, X., Huang, Z., Wang, Z., Wen, C., Jiang, Z., Yu, Z., Liu, J., Liu, G., Huang, X., Maier, A., et~al.: A deep learning based pipeline for optical coherence tomography angiography. Journal of Biophotonics  \textbf{12}(10) (10 2019). \doi{10.1002/jbio.201900008}

\bibitem{multi-gan}
Pan, B., Ji, Z., Chen, Q.: Multigan: Multi-domain image translation from oct to octa. In: Pattern Recognition and Computer Vision. PRCV 2022. Lecture Notes in Computer Science. pp. 336--347. Springer (2022), 13535

\bibitem{diabetic_vessel_loss}
Rosen, R.B., Romo, J.S.A., Krawitz, B.D., Mo, S., Fawzi, A.A., Linderman, R.E., Carroll, J., Pinhas, A., Chui, T.Y.: Earliest evidence of preclinical diabetic retinopathy revealed using optical coherence tomography angiography perfused capillary density. American journal of ophthalmology  \textbf{203},  103--115 (2019). \doi{10.1016/j.ajo.2019.01.012}

\bibitem{song2019first}
Song, G., Chu, K.K., Kim, S., Crose, M., Cox, B., Jelly, E.T., Ulrich, J.N., Wax, A.: First clinical application of low-cost oct. Translational vision science \& technology  \textbf{8}(3),  61--61 (2019)

\bibitem{multi-scale}
Sun, L., Shao, W., Zhu, Q., Wang, M., Li, G., Zhang, D.: Multi-scale multi-hierarchy attention convolutional neural network for fetal brain extraction. Pattern Recognition  \textbf{133},  109029 (2023)

\bibitem{SSIM}
Wang, Z., Bovik, A.C., Sheikh, H.R., Simoncelli, E.P.: Image quality assessment: from error visibility to structural similarity. IEEE Transactions on Image Processing  \textbf{13}(4),  600--612 (2004)

\bibitem{tgu_2d}
Zhang, Z., Ji, Z., Chen, Q., Yuan, S., Fan, W.: Texture-guided u-net for oct-to-octa generation. In: Pattern Recognition and Computer Vision: 4th Chinese Conference, PRCV 2021, Beijing, China, October 29--November 1, 2021, Proceedings, Part IV 4 Springer. pp. 42--52 (2021). \doi{doi.org/10.1007/978-3-030-88013-2_4}

\end{thebibliography}

\end{document}